\documentclass[11pt]{article}
\usepackage[margin=1in]{geometry}
\usepackage{amsmath,amssymb}
\usepackage{graphicx}
\usepackage{booktabs}
\usepackage{longtable}
\usepackage{xcolor}
\usepackage[hidelinks]{hyperref}
\usepackage[numbers,compress,sort]{natbib}
\usepackage{enumitem}
\usepackage{tikz}
\usepackage{placeins}
\usepackage{caption}
\usetikzlibrary{positioning,arrows.meta}

\title{When Can You Trust Neural Critics? Auditing Mutual Information Estimates of Higher-Order Epistasis in Protein Language Models}

\author{Maryam Rahimimovassagh \and Waldemar Karwowski \and Niloofar Yousefi \\
University of Central Florida}
\begin{document}
\maketitle

\begin{abstract}
Mutual information (MI) estimation is a well-established tool for probing what neural
representations encode, and has recently begun to be applied to protein language models to
make order-resolved claims about their encoded structure, but variational MI estimators are
known to have variance that scales at least exponentially in the true MI value, regardless of
estimator family \citep{song2020understanding,mcallester2020formal}. We first audit the
instability's own structure directly: variance across critic seeds within a fixed run and
pairing scheme, matched differences between independently executed runs, and matched
differences between evaluation-pairing schemes on the same trained checkpoint, finding no
systematic bias between runs and a magnitude comparable to ordinary cross-seed retraining
variability, alongside a pairing-scheme sensitivity that independently reproduces the effect
found in Experiment 2's own cross-estimator pairing check. We then test whether an alternative
estimator family resolves the instability at a matched representation dimensionality,
isolating estimator choice from the underlying scientific question: two of three tested
alternatives stabilize the estimate across seeds while one does not. Extending the diagnostic
at the representation dimensionalities of two additional protein language models (a synthetic channel, with no model representations) under blind search, however, all three
estimator families, including both that stabilized at the matched dimension, recover
materially less signal than required, despite correct signal ordering and stable estimates
across seeds. A control critic given the true signal direction, which also reduces its input to a single dimension, recovers $93.3$\% ($5{,}120$-D) and $98.7$\% ($4{,}096$-D) of the analytic MI at gain $1.5$, showing the signal is recoverable with privileged information; because this control changes both what the critic knows and the dimensionality of its input, it does not by itself identify why blind estimation fails. We package the resulting
checks into a reusable diagnostic protocol for order-resolved dependence claims made from
neural critics. Taken together, these results indicate that even a
methodologically stabilized estimator is not currently sufficient for
order-resolved dependence claims made under blind search at this
representation dimensionality and sample size, and that such a claim
should not be reported without the kind of direct stability, pairing,
and signal-recoverability checks this protocol packages.
\end{abstract}

\section{Introduction}

Protein language models (PLMs) are increasingly probed not just for
predictive performance but for what they represent about the structure
of the sequence--function relationship, including epistasis: the
dependence of a mutation's effect on the surrounding sequence context.
Recent work has used PLM outputs to recover higher-order interactions
\citep{tsui2024recovering} and to study the statistics of interacting
sequence motifs learned by these models \citep{zhang2024motifs}, and a
frozen PLM's likelihood scores have independently been shown to align
with experimentally measured epistasis without any explicit MI
estimation step at all \citep{plmepistasis2025}. A parallel line of work,
including our own prior study \citep{rahimimovassagh2026orbitfmib}, has
instead used neural MI estimation to ask a more specific question:
\emph{which interaction orders} of experimentally aligned epistatic
information are retained as a representation propagates through model
depth.

That prior study introduced a fixed-subset protocol that removes a
positional-subset shortcut affecting pooled critics, and used it to ask
whether ESM-2 selectively loses third- and fourth-order epistatic
information relative to first- and second-order information on the
four-site GB1 landscape. An initial production run reported a negative
retention contrast ($\Delta_{\mathrm{HO-LO}} = -0.107$), but an independent replication using identical code, data, and critic seeds reduced this to $-0.017$, with sign changing across otherwise
reasonable evaluation-pairing choices ($-0.011$ to $+0.015$). The
fixed-subset protocol itself held up under replication; the directional
biological claim it produced did not.

This instability is not, on its own, surprising. Neural MI estimators
built on the Donsker--Varadhan representation are known to trade off bias
against variance in ways that are difficult to control in practice
\citep{poole2019variational}, and \citet{song2020understanding} show that
DV- and NWJ-style critics have variance that grows at least
exponentially in the true MI value they are estimating.
\citet{mcallester2020formal} strengthen this into a distribution-free
statement: for a fixed confidence level, no estimator -- not just the
ones tested empirically -- can guarantee accuracy without a sample size
that itself grows exponentially in the quantity being estimated. Recent
work continues to find that the practical reliability of neural MI
estimators in high-dimensional, undersampled regimes is not well
understood, with several 2024--2026 studies proposing new estimators,
new benchmarks, or new methods for flagging untrustworthy estimates
\citep{czyz2023beyond,tuononen2025improving}.

The question this paper asks is therefore not whether neural MI
estimators \emph{can} be unstable, that is established, but whether
the specific instability observed in a fixed-subset protocol applied to
protein language models is consistent with that known theoretical
limitation, whether it is fixable by estimator choice, and whether it persists at the higher representation dimensionalities of other models. We treat this as an audit of a measurement
practice rather than a new estimator paper: our primary output is a
diagnostic protocol other researchers can run before reporting an
order-resolved dependence claim from a neural critic, not a claim that
protein language models do or do not lose higher-order information.

\paragraph{Contributions.}
\begin{enumerate}
    \item An empirical stability analysis of a fixed-subset DV/MINE critic on ESM-2/GB1, separating training variability, run reproducibility, and pairing-scheme sensitivity, each measured directly rather than compared against a theoretical bound that does not, in fact, license a variance claim at this sample size.
    \item A controlled estimator-swap experiment isolating whether the
    instability is a property of the specific estimator used or of the
    measurement regime itself.
    \item A dimension-matched synthetic calibration testing whether the instability (and any estimator fix) persists at the representation dimensionalities of ESM-1v and ProtT5; these calibrations use no model representations.
    \item A released, reusable diagnostic checklist for order-resolved
    dependence claims made from neural critics on frozen representations.
\end{enumerate}

\section{Related Work}

\paragraph{Neural mutual information estimation and its known limits.}
\citet{belghazi2018mine} introduced MINE, a Donsker--Varadhan-based
neural estimator for MI between high-dimensional continuous variables.
\citet{poole2019variational} unify a family of variational MI bounds and
characterize their bias--variance trade-offs. \citet{song2020understanding}
show theoretically and empirically that DV- and NWJ-style critics have
variance that can grow exponentially with the true MI, and propose a
variance-reduction technique based on clipping critic outputs.
\citet{mcallester2020formal} generalize this into a distribution-free
statement about any high-confidence MI lower bound, independent of
estimator architecture. Together, these results imply that any neural
critic applied to a high-dimensional representation with non-trivial
true dependence should be expected to show high variance unless sample
size is large relative to the MI being estimated -- a condition that is
not guaranteed to hold for the small mutational-cube populations
available in protein fitness landscapes.

\paragraph{Benchmarking and diagnosing estimator reliability.}
\citet{czyz2023beyond} systematically benchmark neural and classical MI
estimators across non-Gaussian distributions and find that estimators
significantly underestimate MI outside near-Gaussian regimes. More
recent work continues to probe reliability in high-dimensional,
undersampled settings and to propose numerically stabilized variants
\citep{tuononen2025improving}. \citet{abdelaleem2025accurate} propose an
explicit reliability protocol with confidence intervals for neural MI
estimates and a new class of probabilistic critics for high-dimensional
settings; \citet{eerikinharju2026nmine} extend a MINE-based estimator with
neural marginal-entropy estimation to produce a fully neural normalized MI
estimator; and \citet{foresti2026benchmarks} introduce a copula-theoretic
benchmarking framework spanning synthetic and real-world data and find no
single estimator family dominates across settings, directly consistent
with this paper's own estimator-swap result in Section~\ref{sec:exp2}.

\paragraph{MI estimation and epistasis in protein representations.}
Contrastive, InfoNCE-style MI estimation has previously been formalized
for protein sequence representations \citep{lu2020selfsupervised}, and
provides a natural lower-variance alternative to Donsker--Varadhan-style
critics for the estimator-swap experiment below. Separately from the
MI-estimation route, \citet{plmepistasis2025} show that epistasis
consistent with experimental measurements emerges from PLM likelihood
scores directly, without any neural critic step -- a useful foil for
this paper's argument, since it demonstrates that the same scientific
question can sometimes be asked without inheriting MI-estimator
reliability problems at all.

\paragraph{This work.} Our own prior study \citep{rahimimovassagh2026orbitfmib}
introduced the fixed-subset protocol audited here and reported that its
directional biological finding did not survive independent replication.
This paper treats that non-replication as the object of study: we ask
whether it is explained by known estimator theory, whether it is
estimator-specific, and how to detect it before publication rather than
after.

\section{Audit Protocol}
\subsection{Recap of the Fixed-Subset Measurement}
\label{sec:recap}

\begin{figure}[htbp]
    \centering
    \includegraphics[width=\linewidth]{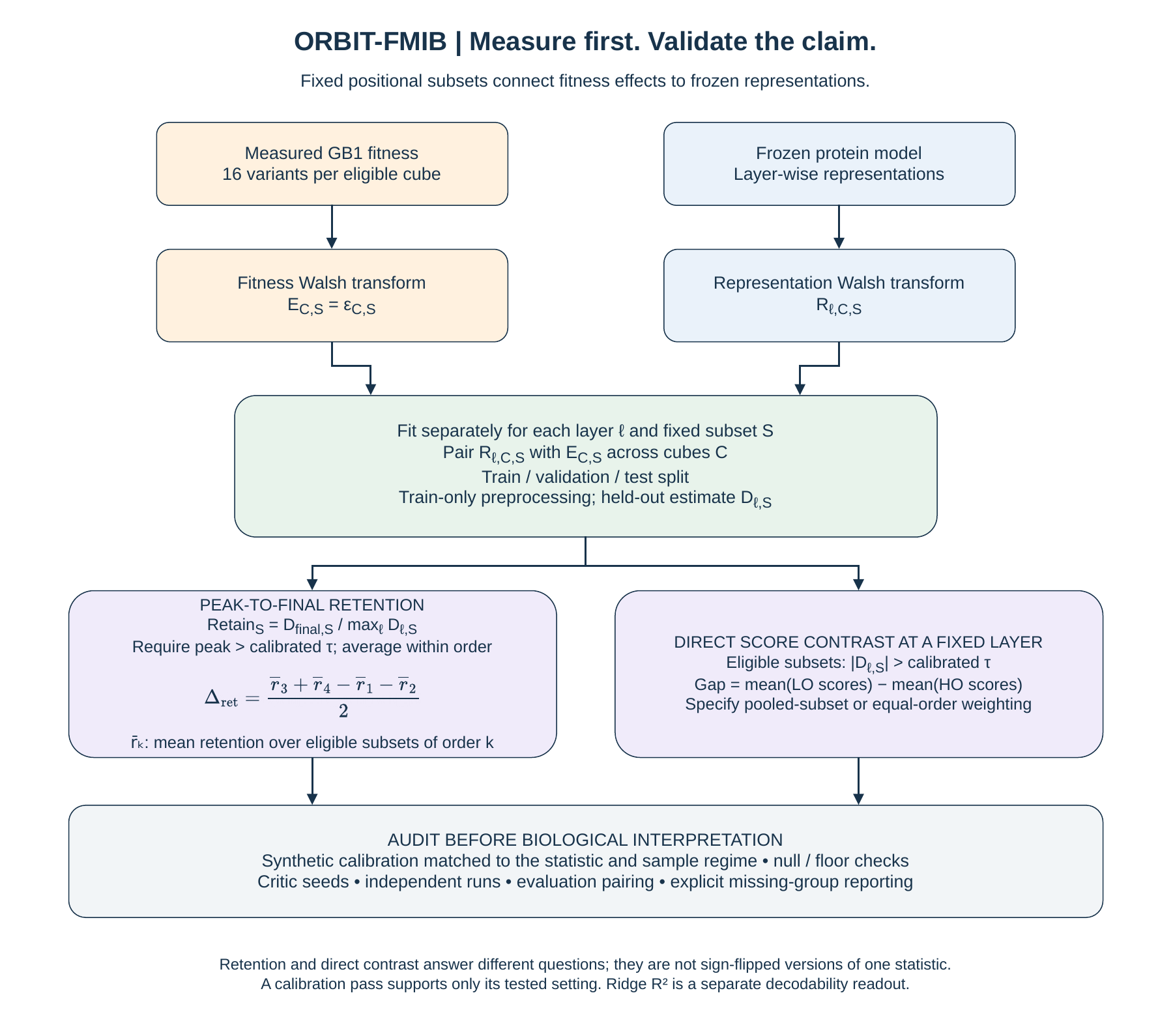}
    \caption{Measurement framework. Fixed positional subsets connect
    experimental fitness effects to frozen model representations via a
    shared Walsh transform, fit separately per layer and subset to
    produce a held-out dependence estimate $D_{\ell,S}$. Two distinct
    statistics are computed from the same $D_{\ell,S}$ grid --
    peak-to-final retention and direct score contrast at a fixed layer
    -- answering different questions rather than being sign-flipped
    versions of one statistic; both require synthetic calibration
    matched to the specific statistic and sample regime before any
    biological interpretation.}
    \label{fig:framework}
\end{figure}

We reuse the $D_{\ell,S}$ estimation step of
\citet{rahimimovassagh2026orbitfmib} without modification for
Experiments 1--3: the same Walsh transform, per-layer/subset critic
fitting procedure, and train/validation/test split. Two distinct
statistics are computed from this shared $D_{\ell,S}$ grid, the
original peak-to-final retention statistic, and a new direct
score-contrast statistic introduced in this paper, answering
different questions rather than being interchangeable or
sign-flipped versions of one another (Figure~\ref{fig:framework}).
Where results below are reported under the new statistic rather than
the original, this is stated explicitly at each point; neither
statistic is treated as a replacement for the other.
Briefly: for each mutational
cube $C$, representation layer $\ell$, and fixed positional subset $S$, a
common Walsh basis yields a representation-level component
$R_{\ell,C,S}$ and an experimental epistatic coefficient $E_{C,S}$.

with $S$ held fixed within each critic to avoid a positional-subset
shortcut. Peak-to-final retention $\mathrm{Retain}_S$ and the order-level
contrast $\Delta_{\mathrm{HO-LO}}$ are computed exactly as in the source
protocol (see \citet{rahimimovassagh2026orbitfmib}, Appendices D--F, for
full definitions). We do not re-derive these here; all new analysis below is additional to, not a replacement for, this protocol.

\paragraph{Data and split.} All analyses use the $149{,}361$ measured four-site GB1 variants of
\citet{wu2016adaptation} and the $109{,}235$ complete wild-type-anchored cubes they contain. The fitness column was
checked variant by variant against the authors' released table (column \texttt{I20fit}, commit \texttt{dbdd7639}): all
$149{,}361$ values agree within $1.1\times10^{-16}$. The ridge results for all three models (Experiment~\ref{sec:exp5}
and Table~\ref{tab:ridge-multimodel}) use the deterministic IID reassignment of the $8{,}210$ identity-strict cubes
($7{,}352/245/613$; split SHA-256 \texttt{4b7f1d17}\ldots); they do not establish identity-disjoint generalization.

\begin{figure}[htbp]
    \centering
    \includegraphics[width=0.85\linewidth]{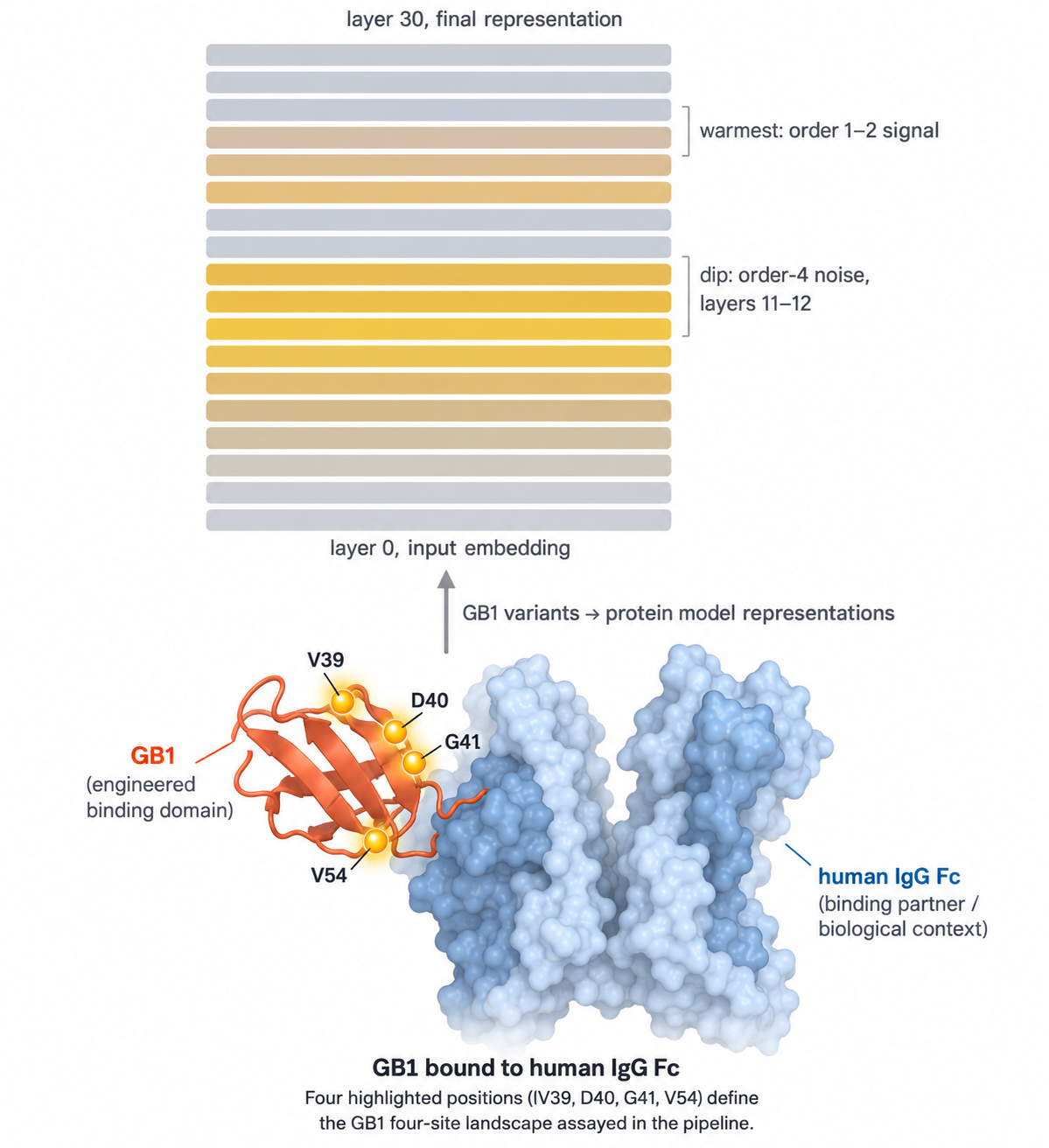}
    \caption{GB1 and its binding partner, the Fc fragment of IgG. GB1
    is a 56-residue domain comprising a four-stranded $\beta$-sheet
    packed against a single $\alpha$-helix; the four positions mutated
    throughout this protocol (39, 40, 41, 54) lie at the binding
    interface with Fc.}
    \label{fig:gb1-structure}
\end{figure}

This departure from a peak-based statistic is deliberate, not incidental.
Reporting the final-to-peak ratio over 31 layers shares a specific structural
feature with this general problem: at each layer, $D_{\ell,S}$ is itself a
noisy estimate, so selecting whichever layer happens to peak inherits the
same selection-induced inflation documented when a maximum is taken over
$N \geq 2$ non-degenerate noisy estimates \citep{singh2025leaderboard}, and
when the largest order statistics of a noisy sample are reported directly
\citep{reid2017postselection}. The layers themselves are not identically
distributed each may carry a genuinely different true signal but the
noise component at each layer is what the peak-selection step is
vulnerable to, independent of that underlying difference. The same selection-induced inflation
has been independently documented in model selection under cross-validation
\citep{mclatchie2024selection}. This motivates testing whether a direct
contrast at a fixed layer, removing the maximum-over-layers step
entirely, resolves the instability, rather than simply reporting
whichever statistic happened to stabilize; the resulting stability
(Table~\ref{tab:exp2}; coefficients of variation dropping one to two
orders of magnitude relative to the original statistic,
Table~\ref{tab:exp1-cov}) is therefore the predicted consequence of
removing this specific source of bias, not an independently convenient
result. COV alone is not treated as decisive here: it can inflate
whenever a statistic's mean approaches zero, exactly the mechanism
Section~\ref{sec:exp1} identifies below for part of the original
statistic's own cross-pairing COV variation. The new statistic's raw
values (Table~\ref{tab:exp2}) stay consistently away from zero across
seeds for every stable estimator, so the low COV here reflects genuine
absolute stability rather than a near-zero-mean artifact. Direct
contrasts of group means, without any peak-normalization
step, are themselves an established convention in representation probing,
most notably mass-mean probing \citep{marks2024geometry}, which computes
class-mean differences directly as a deliberately simpler alternative to
more complex classifiers. Two further checks guard against the new
statistic having been selected post hoc for its stability alone: its
pooled and equal-order-weighted variants agree in sign for every tested
seed, and Ridge regression, a deterministic estimator using neither
this statistic nor this critic family at all
(Section~\ref{sec:exp5}), independently recovers the same
order-dependent pattern.
Figure~\ref{fig:gb1-structure} shows GB1 and its antibody binding
partner, before any combinatorial abstraction is applied to either.

\begin{figure}[htbp]
    \centering
    \includegraphics[width=\linewidth]{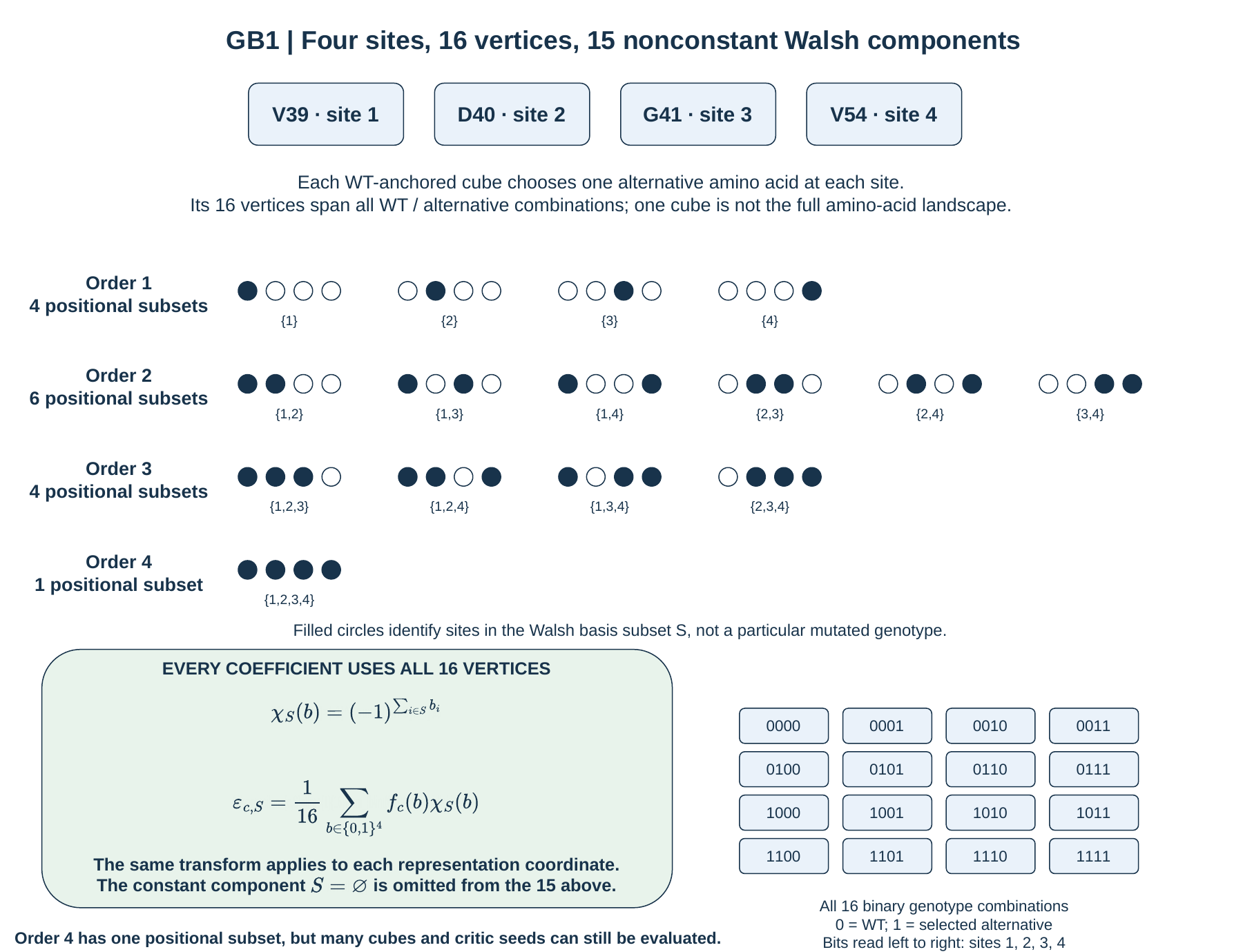}
    \caption{Order-resolved Walsh decomposition of a four-site
    GB1 cube. Each WT-anchored cube selects one alternative amino
    acid at each site and contains all 16 WT/alternative
    combinations. The 15 nonconstant components comprise four
    first-order, six second-order, four third-order, and one
    fourth-order positional subset. Filled circles indicate
    membership in the Walsh basis subset, not mutation states.
    Every coefficient uses all 16 vertices. In the diagram,
    $c$ indexes cubes, $b$ denotes a binary genotype, $f_c(b)$
    denotes its fitness, and $S$ denotes a positional subset.
    The fitness coefficient $\varepsilon_{c,S}$ corresponds to
    $E_{C,S}$ in the main text. The same transform is applied
    coordinate-wise to the representations.}
    \label{fig:gb1-landscape}
\end{figure}

Figure~\ref{fig:gb1-landscape} illustrates the cube construction
and the positional subsets defining the interaction orders.

\subsection{Experiment 1: Empirical Stability Analysis}
\label{sec:exp1}

\textbf{Goal.} Characterize the instability already observed in
$D_{\ell,S}$ directly, along three separable axes: how much it varies
across critic seeds within one run, whether independently executed runs
agree, and how sensitive it is to evaluation-pairing choice on a fixed
trained checkpoint.

\textbf{Method.} We separate three sources of variation that a single
pooled comparison would otherwise conflate. \emph{Training variability}:
for each (layer, subset) cell, the variance of $D_{\ell,S}$ across the
three original critic seeds, holding run and evaluation pairing fixed,
averaged to the interaction-order level. \emph{Run reproducibility}:
matched differences between the original production run and an
independently executed replication, on identical (layer,
subset, seed) cells under the same pairing scheme. \emph{Pairing
sensitivity}: matched differences between evaluation-pairing schemes on
the same trained checkpoint. \citet{mcallester2020formal}'s
distribution-free bound on high-confidence MI lower bounds
($B = O(\log N)$, $\ln N \approx 8.90$ nats at this protocol's sample
size) motivates why this instability is worth auditing at all -- it
constrains achievable point-estimate accuracy at a given sample size,
not variance across retrainings, and is not tested quantitatively here.

To test whether this instability is specifically attributable to the
original statistic's ratio-of-maximum construction, we additionally compute a direct, pooled mean-difference comparison of raw per-layer scores between
the higher- and lower-order groups, with no per-subset maximum and no
ratio (full construction in Appendix~\ref{app:direct-comparison}). We
do not propose this comparison as a general-purpose replacement
estimator; it is constructed narrowly, as a diagnostic to isolate one
specific candidate source of the instability already observed.

\textbf{Result.} The original ratio-based statistic showed cross-seed
coefficients of variation of $0.540$, $14.157$, and $3.240$ under the
cyclic, derandomized, and full-permutation evaluation pairings. The
underlying absolute standard deviations are comparable across all three
($\approx 0.025$--$0.027$); derandomized's large COV reflects its mean
crossing close to zero, not a larger absolute spread in the estimate
itself.
The auxiliary mean-difference comparison showed coefficients of variation
between $0.060$ and $0.342$ across the same pairings, one to two orders
of magnitude smaller throughout, consistent with the ratio-of-maximum
construction itself being a material contributor to the instability,
alongside the sample-size limitation the bound above already predicts.
Run reproducibility was computed as matched differences
(replication $-$ main run) on identical (layer, subset, seed)
cells under cyclic pairing, using two independently written
implementations that share no code, one built specifically to check
the other, rather than reusing its logic, which agree exactly to
every reported digit. Mean signed differences are small and
consistently negative across all four interaction orders ($-0.003$ to
$-0.013$), small relative to both the mean absolute difference
($0.079$--$0.090$) and the standard deviation of the difference
($0.104$--$0.134$), consistent with no systematic bias between runs.
The relevant comparison for magnitude is each run's own cross-seed
standard deviation of the raw dependence score (not the ratio-based
retention statistic, whose cross-seed SD is reported separately
above): this ranges $0.071$--$0.154$ across the two runs and four
orders, placing the run-reproducibility standard deviation in the
same range, consistent with the abstract's ``comparable magnitude''
characterization. Pairing sensitivity, measured as matched
differences (derandomized $-$ cyclic) on the replication run, shows a consistent, substantial positive shift across all four
interaction orders ($0.301$--$0.346$), confirming at the raw-score
level the same pairing-scheme effect Experiment~\ref{sec:exp2} finds
independently for the ratio-based statistic's coefficient of
variation. A cross-check comparing each order's observed
run-reproducibility standard deviation against the value predicted
from the two runs' own cross-seed variances under independence
($\sqrt{v_{\mathrm{main}}+v_{\mathrm{rep}}}$) finds the observed
value consistently at or below prediction (ratio $0.69$--$0.99$
across orders), indicating the two runs' per-cell errors are not
behaving as independent noise, if anything mildly positively
correlated rather than adding unpredictable additional variance.

\paragraph{Execution-environment check (one cell).} For a single cell (layer 9, positional subset mask 3, critic seed 23), the original production run and the independent replication gave test scores of $-0.064$ (best step 725) and $+0.623$ (best step 900). We re-executed identical code, inputs and seed once on a node with a Tesla V100-PCIE-32GB GPU and once on a node with an NVIDIA H100 PCIe GPU (PyTorch 2.7.1+cu126, CUDA 12.6, TF32 matmul disabled; driver version not recorded). The V100 re-execution reproduced the production run's recorded values exactly (same best step, final step and test score to every recorded digit), and the H100 re-execution reproduced the replication's exactly. The GPU models of the two original runs were not recorded, so this is consistent with, but does not establish, that they ran on V100-class and H100-class hardware. The two nodes also differ in CPU and memory configuration, so we do not attribute the difference to the GPU model alone, and this is a single cell and seed. We report it as a caution that identical software and seed need not yield identical critic output across execution environments, not as a general result about hardware sensitivity.
\begin{table}[h]
\centering
\caption{Retraining stability by pairing scheme (Experiment~1). Coefficient of variation (COV) of $\Delta_{\mathrm{HO-LO}}$ across the 3 critic-retraining seeds, for the old (peak-ratio) statistic and the new (direct-comparison) statistic under both weighting conventions. Lower COV indicates greater stability across independent retrains of the same critic.}
\label{tab:exp1-cov}
\begin{tabular}{lccc}
\toprule
Pairing scheme & Old statistic COV & New (pooled) COV & New (equal-weighted) COV \\
\midrule
Cyclic & $0.540$ & $0.272$ & $0.161$ \\
Derandomized & $14.157$ & $0.152$ & $0.060$ \\
Full permutation & $3.240$ & $0.342$ & $0.252$ \\
\bottomrule
\end{tabular}
\end{table}

\begin{table}[h]
\centering
\caption{Training variability: mean cross-seed variance per (layer,
subset) cell, by interaction order $k$, computed separately for each
run. Both runs use the standard cyclic-shift evaluation.}
\begin{tabular}{ccc}
\toprule
$k$ & Main run & replication \\
\midrule
1 & 0.0060 & 0.0051 \\
2 & 0.0148 & 0.0138 \\
3 & 0.0101 & 0.0141 \\
4 & 0.0238 & 0.0140 \\
\bottomrule
\end{tabular}
\end{table}

\begin{table}[h]
\centering
\caption{Pairing sensitivity: matched differences (derandomized $-$
cyclic) on the replication run, by interaction order $k$.}
\begin{tabular}{ccccc}
\toprule
$k$ & $n$ matched & Mean diff. & Mean $|$diff.$|$ & SD of diff. \\
\midrule
1 & 372 & 0.322 & 0.322 & 0.091 \\
2 & 558 & 0.333 & 0.334 & 0.148 \\
3 & 372 & 0.346 & 0.346 & 0.155 \\
4 & 93 & 0.301 & 0.301 & 0.124 \\
\bottomrule
\end{tabular}
\end{table}

\begin{table}[h]
\centering
\caption{Run reproducibility: matched differences (replication $-$ main), cyclic scheme, by interaction order $k$.}
\begin{tabular}{ccccc}
\toprule
$k$ & $n$ matched & Mean diff. & Mean $|$diff.$|$ & SD of diff. \\
\midrule
1 & 372 & -0.003 & 0.079 & 0.104 \\
2 & 558 & -0.005 & 0.081 & 0.121 \\
3 & 372 & -0.013 & 0.082 & 0.127 \\
4 & 93 & -0.006 & 0.090 & 0.134 \\
\bottomrule
\end{tabular}
\end{table}

Cross-check (predicted vs.\ observed cross-run SD from the two
within-run variances, $\sqrt{v_{\mathrm{main}}+v_{\mathrm{rep}}}$):
$k=1$: predicted 0.106, observed 0.104; $k=2$: predicted 0.169, observed 0.121; $k=3$: predicted 0.156, observed 0.127; $k=4$: predicted 0.195, observed 0.134.

\subsection{Experiment 2: Estimator Swap (ESM-2 only)}
\label{sec:exp2}

\textbf{Goal.} Isolate whether the instability is fixable by estimator
choice, holding the model, landscape, and cube partition fixed.

\textbf{Method.} We compare the original Donsker--Varadhan/MINE critic
against: (a) an InfoNCE/CPC-style contrastive estimator, following the
formalization used for protein sequence representations in
\citet{lu2020selfsupervised}; (b) the SMILE critic of
\citet{song2020understanding}, in which only the product-of-marginals
term of the Donsker--Varadhan bound is clipped to
$[-\tau_{\mathrm{clip}}, \tau_{\mathrm{clip}}]$ before the
log-partition term; the critic is trained via the Jensen--Shannon
objective their own reference implementation uses, not by directly
maximizing the clipped bound itself\footnote{Directly maximizing the
clipped bound is unbounded as a training objective: a critic that
ignores its inputs and outputs an increasing constant achieves an
unboundedly large score, since the joint term is deliberately left
unclipped in the true formula.}; and (c) a symmetric-clipped bounded-DV
variant, in which both the joint and product-of-marginals terms are
clipped. This differs from (b) but represents a related, independently
reasonable point on the same bias--variance trade-off, and is included
alongside it as a natural comparison. All estimators use the same 15
fixed positional subsets, the same 31 representation stages, the same
deterministic IID cube partition, and the same critic seeds
$(11, 23, 37)$ as the original protocol. We recompute
$\Delta_{\mathrm{HO-LO}}$ under each estimator and test cross-seed,
cross-run, and cross-evaluation-pairing stability exactly as in the
original replication study.

\textbf{Decision rule.} If any candidate estimator stabilizes
$\Delta_{\mathrm{HO-LO}}$ across seeds, independent runs, and evaluation pairings
where DV/MINE did not, we adopt it as the recommended estimator for this
class of analysis and rerun the original biological question under it.
If no candidate stabilizes the estimate, this strengthens the claim that
no neural critic is currently safe to use for this class of directional
claim at this sample size and dimensionality.

\textbf{Result.} Each candidate estimator was calibrated against its
own null-derangement floor $\tau$ (a derangement is a permutation with
no fixed points, so no null draw ever pairs an item with itself)
rather than reusing DV/MINE's; the
four resulting values span roughly a fivefold range, from
$\tau = 0.007634$ for DV/MINE\footnote{This is the value from the IID null. Some run manifests, including the
comparison summaries behind the direct-contrast statistic, record $\tau = 0.006777$ for DV/MINE, taken from an earlier
strict-partition null. We did not test whether this choice changes the downstream statistics.} to $\tau = 0.038056$ for InfoNCE.
Table~\ref{tab:exp2} reports $\Delta_{\mathrm{HO-LO}}$ under the
direct, pooled mean-difference comparison of Section~\ref{sec:exp1} for each
candidate across the three original critic seeds, alongside DV/MINE's
own value under the same evaluation-pairing convention (cyclic) for
direct comparison. InfoNCE does not stabilize the estimate: its sign
changes across all three seeds ($-0.116$, $+0.041$, $+0.091$). DV/MINE,
SMILE, and the bounded-DV variant are each sign-consistent across all
three seeds under this single pairing convention ($+0.143$ to $+0.284$;
$+0.172$ to $+0.235$; $+0.134$ to $+0.204$, respectively). Cross-run
replication now exists for two of the three candidates: Bounded-DV and
SMILE's pairing-sensitivity reruns (Table~\ref{tab:pairing-sensitivity})
were independently re-executed against their original grids, and
both are sign-stable across all three seeds against that original grid,
matching DV/MINE's own cross-run stability from
Section~\ref{sec:exp1}. InfoNCE has no independent rerun yet.

\textbf{Decision rule applied.} On cross-seed stability under a single
evaluation-pairing convention, InfoNCE is the sole candidate that fails
to stabilize; DV/MINE itself is already stable by this specific
criterion, so this comparison alone does not show SMILE or the
bounded-DV variant \emph{fixing} an instability DV/MINE has under
cyclic pairing specifically. The instability motivating this experiment
is more precisely a cross-evaluation-pairing effect (Table~\ref{tab:exp1-cov},
Section~\ref{sec:exp1}): DV/MINE's cross-seed COV under cyclic pairing
alone is $0.540$, but reaches $14.157$ under derandomized pairing.
Cross-pairing stability has now been measured at full-grid scale for
DV/MINE, SMILE, and the bounded-DV variant alike (Section~\ref{sec:exp4-pairing}), and all three are also cross-run stable. The decision rule as
stated, requiring cross-seed, cross-run, \emph{and} cross-pairing
stability, can therefore now be fully evaluated for these three;
InfoNCE remains the one candidate for which it cannot, having failed
cross-seed stability outright and having no cross-run run at all.
This instability under the original statistic is not unique to
DV/MINE: SMILE's own old-statistic values sign-flip across the same
three seeds ($-0.043$, $-0.032$, $+0.010$), the identical pattern
motivating this paper's departure from that statistic in the first
place (Section~\ref{sec:recap}), now independently reproduced in a
second estimator rather than resting on DV/MINE's case alone.
\textbf{ORBIT-Synth validation.} The full ORBIT-Synth validation suite
(summarized in Appendix~\ref{app:orbit-synth}; full detail in
Appendices H.1--H.3 of \citet{rahimimovassagh2026orbitfmib}:
known-ground-truth recovery, dimension-specific absolute-MI calibration,
relative-retention check) was run for InfoNCE, SMILE, and the bounded-DV
variant, matching what DV/MINE underwent. All three pass every
acceptance gate (Table~\ref{tab:orbit-synth}). This rules out the
concern motivating this check for all three: an estimator merely
unresponsive to $E_{C,S}$ would still fail to recover the strong,
known synthetic signal this suite tests against, and none did. For
InfoNCE specifically, this sharpens rather than resolves its
instability: it correctly recovers a strong synthetic signal with low
cross-seed variance (SD $0.0305$, well inside the $0.08$ threshold),
yet still shows sign-flipping cross-seed instability on the real,
weaker GB1 signal (Section~\ref{sec:exp2}). Its real-data instability
is therefore not an artifact of general insensitivity to signal; it is
a property specific to the weaker, noisier regime the real biological
question sits in.

\begin{table}[h]
\centering
\caption{ORBIT-Synth validation results, all three alternative estimators (Experiment 2). Columns: maximum cross-seed standard deviation; the synthetic $\Delta_{\mathrm{HO-LO}}$ selectivity contrast; the minimum recovered estimate on the strong synthetic channel; the maximum $|$estimate$|$ on the null-control channel. All four acceptance gates pass for each estimator; full gate-by-gate detail (Walsh reconstruction, bottleneck localization, order-level signal-minus-null) is in each estimator's archived validation manifest.}
\label{tab:orbit-synth}
\begin{tabular}{lcccc}
\toprule
Estimator & Max SD & $\Delta_{\mathrm{HO-LO}}$ & Strong min.\ & Null max.\ \\
\midrule
InfoNCE & $0.0305$ & $-0.9378$ & $0.5319$ & $0.0014$ \\
Bounded-DV & $0.0087$ & $-0.9705$ & $0.5278$ & $0.0022$ \\
SMILE & $0.0070$ & $-0.9641$ & $0.5264$ & $0.0023$ \\
\bottomrule
\end{tabular}
\end{table}

\textbf{Instability characterization.} Three descriptive checks probe
what kind of instability is present, without identifying its cause.
First, for the bounded-DV variant, cross-seed noise shows little
observed linear association with product-term clip activation
(correlation $0.007$ against COV, $0.077$ against raw SD, $n=450$
cells) -- clipping is active in 97.4\% of fits, but the degree of
activation does not predict where this estimator is noisier. Second,
cross-seed SD broken out by interaction order reveals a four-way split:
Bounded-DV, SMILE, and InfoNCE each show lower noise at higher order,
while DV/MINE shows the opposite, noisiest at order 4 and least noisy
at order 1. Third, for DV/MINE specifically, cross-seed noise is
negatively correlated with observed signal magnitude ($-0.409$ for
$|\mathrm{mean}(x)|$, $-0.369$ for $\mathrm{mean}(|x|)$) -- noisiest
where the observed effect is weakest, not strongest. This is not a
direct test of \citet{song2020understanding}'s variance-grows-with-true-MI
result, since this protocol makes no claim to calibrated absolute MI;
it is, at most, evidence that a naive reading of that result does not
describe what is observed here. All three findings are descriptive and
do not establish what drives the instability in question.

\paragraph{Order-resolved depth curves.} The pooled contrast above
summarizes each estimator's higher-order-versus-lower-order gap as a
single number at one layer; extracting the full per-order,
per-(layer,estimator) grid shows this pattern held consistently across
nearly the entire depth, not only at the pooled summary point
(Table~\ref{tab:depth-curves-layer30}; full depth curves in
Appendix~\ref{app:depth-curves}, Figure~\ref{fig:depth-curves} and
Figure~\ref{fig:depth-curves-by-order}). Two patterns are apparent.
First, SMILE reports systematically higher scores than DV/MINE and
Bounded-DV across every order and nearly the full depth, while DV/MINE
and Bounded-DV track much closer to one another; consistent with the
asymmetric interpretation already applied to Ridge in
Section~\ref{sec:exp5}, this is not treated as evidence SMILE recovers
more true information, since no ground-truth MI value is available on
this real biological landscape to calibrate any estimator's absolute
scale against. Second, DV/MINE's order-4 curve shows a pronounced dip
around layers 11--12 not present at the same layers for Bounded-DV or
SMILE -- direct visual confirmation of the cross-seed noise
concentration at order 4 already reported above for DV/MINE
specifically, rather than a new, independent finding.

\begin{table}[h]
\centering
\caption{Mean held-out MI estimate, in nats, by interaction order at layer 30, all three estimators, extracted directly from the same validated grid Figure~\ref{fig:depth-curves} and Figure~\ref{fig:depth-curves-by-order} (Appendix~\ref{app:depth-curves}) are plotted from, at the specific layer Table~\ref{tab:exp2} also uses. Each cell averages seeds within subset first, then subsets within order (12 seed-subset cells for orders 1 and 3, 18 for order 2, 3 for order 4). No threshold filtering is applied.}
\label{tab:depth-curves-layer30}
\begin{tabular}{lcccc}
\toprule
Estimator & Order 1 & Order 2 & Order 3 & Order 4 \\
\midrule
DV/MINE    & 0.9953 & 0.8820 & 0.7284 & 0.6622 \\
Bounded-DV & 0.9460 & 0.8845 & 0.7552 & 0.6749 \\
SMILE      & 1.2097 & 1.0792 & 0.9449 & 0.8233 \\
\bottomrule
\end{tabular}
\end{table}

\FloatBarrier

\subsection{Experiment 3: Dimension-Matched Synthetic Calibration}
\label{sec:exp3}

\textbf{Goal.} Test whether the estimator behavior found on ESM-2 persists at the higher representation
dimensionalities of two other protein language models, on a synthetic channel with known mutual information.

\textbf{Method.} We run the synthetic strong-signal calibration of the source protocol at the representation
dimensionalities of ESM-1v ($5{,}120$-D) and ProtT5 ($4{,}096$-D): a Gaussian channel with known mutual information,
$R = g\,E\,u + \sigma\eta$ with $E\sim\mathcal N(0,1)$, $\|u\|=1$ and $\sigma=1$, at the GB1 sample regime
($7{,}352$ training, $245$ validation and $613$ test cubes), using the three estimators of
Experiment~\ref{sec:exp2}. \emph{No ESM-1v or ProtT5 representations are used}: the models enter only through input
dimension and critic width. The original DV/MINE critic failed absolute-MI calibration at these dimensionalities in
\citet{rahimimovassagh2026orbitfmib}.

\textbf{Decision rule.} Consistent failure at both dimensionalities supports a statement about estimator
reliability at that dimensionality and sample size, independent of any model; it does not support claims about the
models themselves.

\textbf{Result.} On ESM-2, the original DV/MINE baseline passes calibration
with $\Delta_{\mathrm{HO-LO}} = -0.107$, though
Experiment~\ref{sec:exp1} already establishes this specific value as
unstable under replication. At the $5{,}120$-D and $4{,}096$-D calibration channels, all three estimators tested in
Experiment~\ref{sec:exp2} -- DV/MINE, Bounded-DV, and SMILE -- fail
the same acceptance gate (strong-signal margin) at both dimensionalities,
despite correct signal ordering and stable seed-level estimates for
every attempt; full replicate-level detail is in
Appendix~\ref{sec:margin-gate-appendix}. A control critic given the true signal direction, which also reduces its
input to one dimension, recovers the signal at both dimensionalities: its mean estimate is $93.3$\% ($5{,}120$-D) and
$98.7$\% ($4{,}096$-D) of the analytic MI
at gain $1.5$, and $108$\% and $94$\% at gain $1.0$
(Appendix~\ref{sec:margin-gate-appendix}). The signal is therefore
recoverable with privileged information at this representation
dimensionality; because the control changes both the critic's knowledge
and its input dimensionality, it does not identify which of the two
accounts for the blind estimators' failure.

A separate check, on the models' own representations: ridge regression (Section~\ref{sec:exp5}), which involves no training or
search and is therefore immune by construction to the retraining
instability characterized in Section~\ref{sec:exp1}, recovers the same
monotonic order-dependent pattern on both ESM-1v and ProtT5
(Table~\ref{tab:ridge-multimodel}): held-out $R^2$ declines with
interaction order on both models (ESM-1v, layers 1--33: order 1,
$0.991$; order 2, $0.974$; order 3, $0.934$; order 4, $0.853$; ProtT5,
layers 1--24: order 1, $0.983$; order 2, $0.961$; order 3, $0.919$;
order 4, $0.846$), directionally consistent with the ESM-2 ridge result (Section~\ref{sec:exp5}). This is the only evidence in this
paper drawn from these two models' own representations: it indicates that linearly decodable order-dependent structure
is present in them under the IID split; the synthetic calibration above does not bear on that question. Because higher-order coefficients may be noisier targets by
construction, part of the decline in $R^2$ with order may reflect target
noise rather than representation content; we do not yet report $R^2$
against a per-order reliability ceiling. A robustness audit of these two new grids, matching the one
already applied to ESM-2, confirms numerical stability and a
cleanly collapsing null for both models (Section~\ref{sec:discussion}).


\begin{table}[h]
\centering
\caption{Ridge mean held-out $R^2$ ($D_{\ell,S}$) by interaction order,
ESM-1v and ProtT5. Layer 0 is reported separately, matching the paper's
own ESM-2 convention; the main summary covers layers 1--33 (ESM-1v) /
1--24 (ProtT5). Mean and SD for the layers-1-$N$ rows are computed
across each layer's own per-order mean (layer-to-layer variation), not
across individual layer--subset cells. The LO/HO contrast weights each
interaction order equally (mean of the two order-level means), matching
the paper's existing ESM-2 convention, not a pooled mean over all
cells.}
\label{tab:ridge-multimodel}
\begin{tabular}{ccccc}
\toprule
$k$ & ESM-1v layer 0 & ESM-1v layers 1--33 (SD) & ProtT5 layer 0 & ProtT5 layers 1--24 (SD) \\
\midrule
1 & 0.4773 & 0.9911 (0.0017) & 0.4774 & 0.9834 (0.0013) \\
2 & -0.0015 & 0.9738 (0.0023) & -0.0015 & 0.9608 (0.0040) \\
3 & -0.0009 & 0.9339 (0.0057) & -0.0009 & 0.9186 (0.0057) \\
4 & -0.0000 & 0.8531 (0.0154) & -0.0000 & 0.8464 (0.0140) \\
\midrule
LO (1,2) mean & -- & 0.9824 & -- & 0.9721 \\
HO (3,4) mean & -- & 0.8935 & -- & 0.8825 \\
\bottomrule
\end{tabular}
\end{table}

\subsection{Experiment 4: Diagnostic Protocol}
\label{sec:exp4}

Based on Experiments~\ref{sec:exp1}--\ref{sec:exp3}, we package the
following checklist for order-resolved dependence claims made from
neural critics on frozen representations:

\begin{enumerate}[label=(\arabic*)]
    \item Compute a floor threshold from a cube-derangement (or
    equivalent) null before interpreting any dependence score.
    \item Run a positional-subset (or equivalent structural) shortcut
    control comparing pooled and fixed-subset critics.
    \item Measure stability directly along three axes, cross-seed training variability, cross-run reproducibility, and cross-pairing sensitivity, rather than against a theoretical bound that does not license a variance claim (Experiment~\ref{sec:exp1}).
    \item Require at least three independent seed-and-run replications before reporting a directional claim.
    \item Test at least two evaluation-pairing schemes on the same
    trained critic.
    \item Do not report a directional interaction-order claim unless it
    survives all of the above.
\end{enumerate}

Code implementing steps 1--5 as automatic checks is available from the authors on request.

\subsection{Cross-Estimator Pairing-Sensitivity Check}
\label{sec:exp4-pairing}

Checklist item (5) above calls for testing at least two evaluation-pairing
schemes on the same trained critic. We ran this check over the full
measurement grid for all three candidate estimators (31 layers $\times$
15 fixed subsets $\times$ 3 seeds, layer 0 excluded as a floor case where
the ratio is unstable near zero signal; $n=1350$ per estimator). A
literal port to InfoNCE does not apply, since InfoNCE's bound is provably
invariant to joint-side permutation of the full test set.

Table~\ref{tab:pairing-sensitivity} reports the ratio of each critic's
standard cyclic-shift evaluation to its own estimate under two
alternative negative-pairing schemes. All three estimators fall within a
five-percentage-point band (69.7\%--74.3\%), despite differing in
clipping and training objective: DV/MINE at 69.7\%/70.1\%, the
bounded-DV variant at 71.7\%/71.8\%, and SMILE at 73.8\%/74.3\%. This
convergence across three Donsker--Varadhan-based estimators that differ
in clipping and training objective, now confirmed at full-grid scale for
all three, is consistent with cyclic-shift pairing carrying a systematic
underestimate of roughly this magnitude that does not depend on those
differences; we did not test whether it extends beyond this estimator
family.

\begin{table}[h]
\centering
\caption{Cyclic-shift pairing sensitivity, by estimator. Ratio of the
standard cyclic-shift \texttt{test\_mi} to the same critic's estimate
under two alternative negative-pairing schemes. Reported statistic is
the MEAN of the per-fit ratio (5th--95th percentile range shown in
parentheses); $n$ is fits actually used per comparison after excluding
layer 0 and any non-finite or near-zero-denominator cases.}
\label{tab:pairing-sensitivity}
\begin{tabular}{lccc}
\toprule
Estimator & $n$ (cyc/derand, cyc/fullperm) & Cyclic / derandomized & Cyclic / full-permutation \\
\midrule
DV/MINE & 1350, 1350 & 69.7\% (53.5\%--83.1\%) & 70.1\% (51.1\%--87.9\%) \\
Bounded-DV & 1350, 1350 & 71.7\% (58.2\%--83.2\%) & 71.8\% (57.4\%--85.5\%) \\
SMILE & 1350, 1350 & 73.8\% (62.7\%--83.9\%) & 74.3\% (62.3\%--88.2\%) \\
\bottomrule
\end{tabular}
\end{table}

\subsection{Experiment 5: Ridge Regression as a Structurally Independent Robustness Check}
\label{sec:exp5}

\textbf{Goal.} Test whether a deterministic, closed-form estimator --
immune by construction to the retraining variance driving this audit --
detects a directionally consistent signal at the same (layer, subset)
cells the trained critics are evaluated on.

\textbf{Method.} For each (layer, subset) cell, we fit ridge regression
of $E_{C,S}$ on $R_{\ell,C,S}$ with regularization strength selected by
held-out validation $R^2$, reporting test-set $R^2$ as $D_{\ell,S}$.
Given a fixed train/validation/test split, this is a closed-form
solve with no stochastic training step: unlike DV/MINE, for which only
318 of 1{,}395 replicated fits (23\%) selected the same best-validation
step as the original run (mean absolute held-out difference 0.081 nats,
max 0.688), ridge's result given identical data is bit-identical by
construction, not merely empirically stable. A derangement-null variant
(every target guaranteed paired with a different example, not merely
permuted) establishes the floor $\tau$ below which no signal is
distinguishable from chance.

\textbf{Interpretation rule, stated asymmetrically.} Ridge $R^2$
measures linear decodability only; DV/MINE and its alternatives estimate
nonlinear mutual information, reported in nats. These are not the same
statistical quantity and are not reported in Table~\ref{tab:exp2} or
compared via a shared $\tau$. A high ridge $D_{\ell,S}$, well above its
own derangement-null floor, is informative on its own: corroborating
evidence of real, linearly-decodable dependence from an estimator
structurally immune to the instability characterized in
Section~\ref{sec:exp1}. A low or near-zero ridge $D_{\ell,S}$ is
\emph{not} evidence of no dependence -- it is equally consistent with
real dependence that is simply nonlinear, and is not treated as
disconfirming any trained-critic result.

\begin{figure}[h]
\centering
\includegraphics[width=0.75\textwidth]{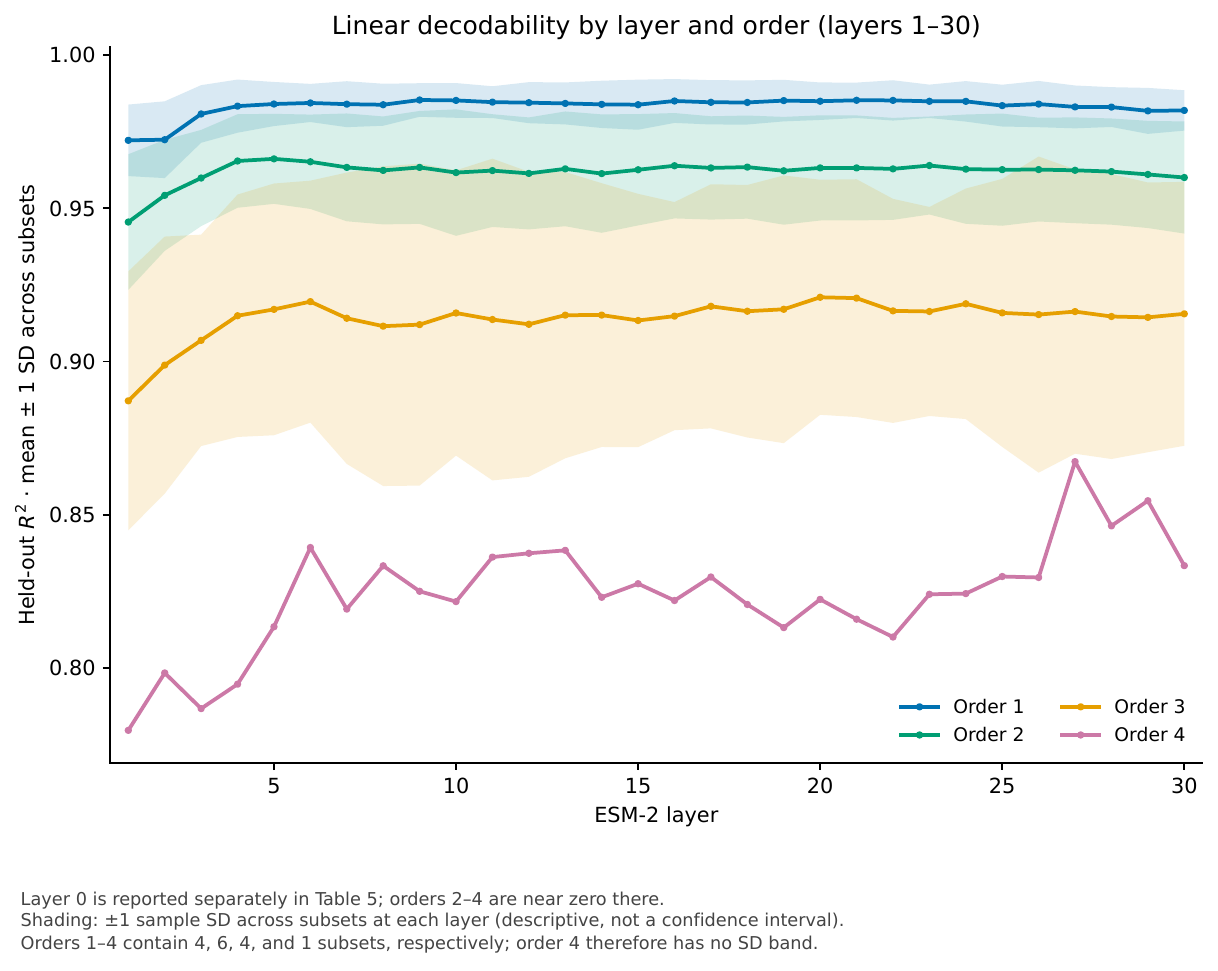}
\caption{Linear decodability by layer and interaction order, layers 1--30. Layer 0 is a floor case (near zero for orders 2--4) reported separately in Table~\ref{tab:exp5-ridge}. Shading: $\pm$1 sample SD across subsets at each layer (descriptive, not a confidence interval).}
\label{fig:exp5-ridge}
\end{figure}

\textbf{Result.} Table~\ref{tab:exp5-ridge} and Figure~\ref{fig:exp5-ridge}
report the full 31-layer, 15-subset grid -- a snapshot comparison
across layers, not a claim that information is progressively lost
with depth. At layer 0, mean held-out $R^2$ is near zero for orders
2--4 ($|D_{\ell,S}| < 0.003$), indicating little linear decodability
under this probe and split, while order 1 carries appreciable signal
($D_{\ell,S}=0.48$) even at this layer -- consistent with, though not
independently establishing, DV/MINE's own layer-0 result. Across
layers 1--30, held-out $R^2$ declines monotonically with interaction
order (order 1: $0.983$; order 2: $0.962$; order 3: $0.914$; order 4:
$0.824$), directionally consistent with the Bounded-DV/SMILE pattern
from a structurally independent, deterministic estimator. Averaging
each order's own mean equally, low-order (1,2) vs.\ high-order (3,4)
contrast is $0.9726$ vs.\ $0.8689$; neither this contrast nor the
per-order means above is itself a retention statistic. A targeted
robustness audit (Section~\ref{sec:discussion}) confirms this result is
numerically stable and that a properly derived null collapses
correctly.

\begin{table}[h]
\centering
\caption{Experiment 5 (Ridge regression robustness check). $D_{\ell,S}$ is held-out $R^2$ (linear decodability), a different statistical quantity from the MI estimates DV/MINE and its alternatives report in nats. Layer 0 column: mean $D_{\ell,S}$ at the raw embedding layer. Layers 1--30 columns: mean and SD across the 30 per-layer means, subsets averaged within each layer first.}
\label{tab:exp5-ridge}
\begin{tabular}{lccc}
\toprule
Order $k$ & Layer 0 & Mean, layers 1--30 & SD, layers 1--30 \\
\midrule
1 & $0.4773$ & $0.9833$ & $0.0032$ \\
2 & $-0.0015$ & $0.9619$ & $0.0037$ \\
3 & $-0.0009$ & $0.9140$ & $0.0065$ \\
4 & $-0.0000$ & $0.8239$ & $0.0184$ \\
\bottomrule
\end{tabular}
\end{table}

\section{Results}

Table~\ref{tab:exp2} reports Experiment 2's results in full.
Experiment 1's empirical stability results are reported in
Section~\ref{sec:exp1}. Experiment 3's results, including the
dimension-matched calibration-outcome table, are reported in
Section~\ref{sec:exp3}.

\begin{table}[h]
\centering
\caption{$\Delta_{\mathrm{HO-LO}}$ stability across estimators (Experiment 2), ESM-2 only. Values are the pooled raw-score mean difference (lower-order minus higher-order), evaluated at layer 30, from Section~\ref{sec:exp1}'s direct-comparison diagnostic, not the original ratio-based statistic; the sign convention is therefore reversed relative to $\Delta_{\mathrm{HO-LO}}$ itself, so a positive value here indicates lower-order exceeding higher-order, the same direction as the original finding. DV/MINE row uses the cyclic evaluation-pairing scheme, matching the single convention the other three estimators' main grids were run under. Cross-run entries report sign agreement (matched seed pairs) between each estimator's original grid and an independent second run; for Bounded-DV and SMILE this reuses their pairing-sensitivity rerun (Table~\ref{tab:pairing-sensitivity}) rather than a dedicated rerun.}
\label{tab:exp2}
\begin{tabular}{lcccc}
\toprule
Estimator & Seed 1 (11) & Seed 2 (23) & Seed 3 (37) & Cross-run \\
\midrule
DV/MINE (baseline) & $0.209$ & $0.143$ & $0.284$ & Stable (3/3) \\
InfoNCE & $-0.116$ & $0.041$ & $0.091$ & Sign-agree (3/3)$^\dagger$ \\
Bounded-DV & $0.204$ & $0.134$ & $0.172$ & Stable (3/3) \\
SMILE & $0.225$ & $0.235$ & $0.172$ & Stable (3/3) \\
\bottomrule
\end{tabular}
\end{table}
InfoNCE's cross-run agreement is real (all three matched seeds agree in
sign between its original grid and its rerun) but does not indicate reliability:
InfoNCE already fails cross-seed stability in each run individually
(sign flips across seeds 11/23/37, this table's own Seed~1--3 columns), and
its cross-seed COV reaches $16.740$ in the original grid and $5.382$ in
the rerun -- both comparable in scale to DV/MINE's own worst-case
instability under derandomized pairing ($14.157$, Table~\ref{tab:exp1-cov}).
Cross-run sign agreement and cross-seed stability are different,
independent checks; InfoNCE passes the former and fails the latter.

\begin{figure}[h]
\centering
\includegraphics[width=\textwidth]{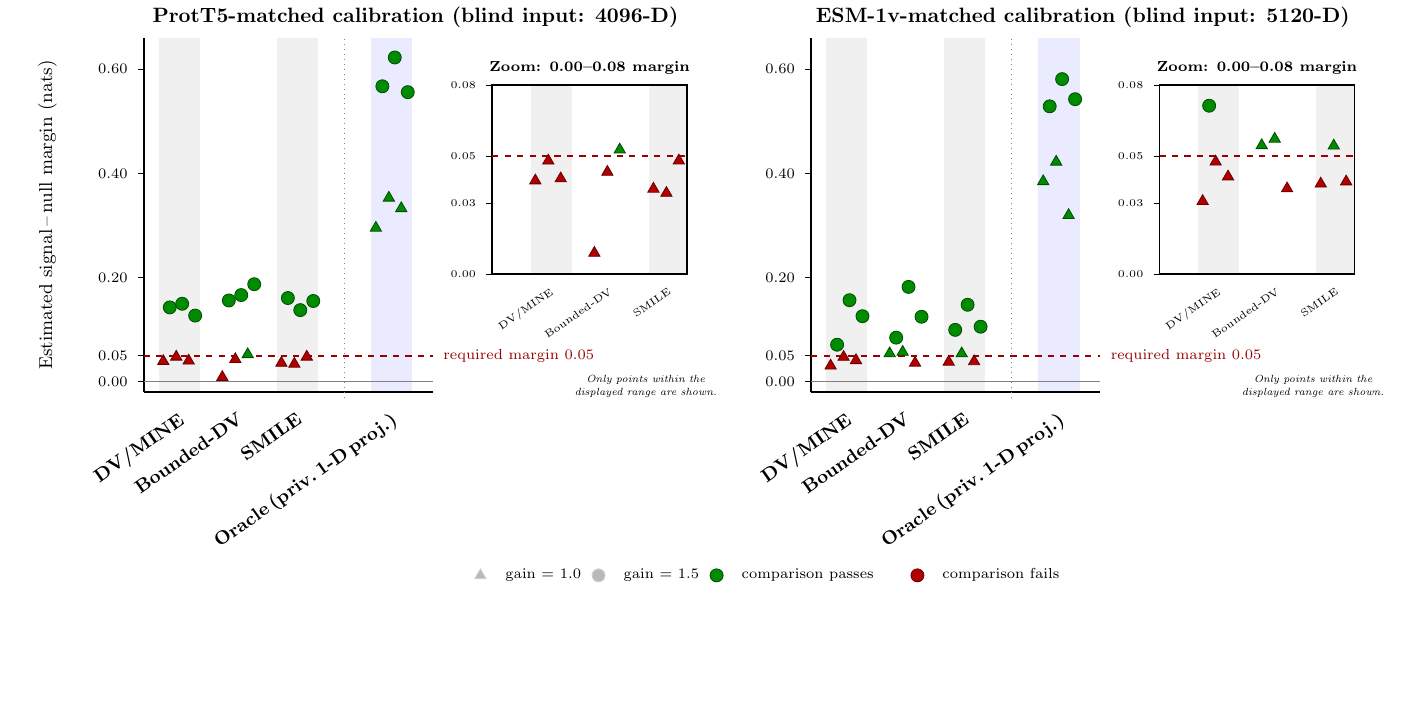}
\input{figures/fig_margin_gate_caption}
\end{figure}

\FloatBarrier

\section{Discussion}
\label{sec:discussion}
A structurally independent check corroborates the pattern above, and
survives a targeted robustness audit. A 45-cell spot-check (layers 1,
15, and 30, all subsets) refit with float64 precision and a
numerically stable solver reproduces the original scores closely
(max absolute difference $0.0015$, mean $0.0001$), confirming the
ill-conditioned-matrix warnings observed during fitting did not
materially affect the reported values. A properly within-split
derangement null (five seeds, full regularization reselection per
draw) collapses correctly: no null draw approached the real signal
range across any of the 225 fits (max $D_{\mathrm{null}} = 0.0056$,
against real scores of $0.78$--$0.99$), ruling out the most direct
form of target-handling error. This audit tests numerical robustness
and sensitivity to genuine feature-target pairing under the existing
IID split; it does not establish identity-disjoint generalization,
which was not tested.

We audited 45 cells per model and repeated each null fit with five derangement seeds. The maximum and mean absolute differences between the original and float64/SVD scores were $0.0038$ (max) and $0.0003$ (mean) for ESM-1v and $0.0007$ (max) and $0.0000$ (mean) for ProtT5. Across 225 null fits per model, the maximum null scores were $-0.0024$ (ESM-1v) and $-0.0066$ (ProtT5), compared with observed score ranges of $0.8398$--$0.9942$ (ESM-1v) and $0.8027$--$0.9888$ (ProtT5). These measurements concern the audited cells under the existing IID split; they do not establish numerical stability throughout the full grids or identity-disjoint generalization.

Taken together, Experiments~\ref{sec:exp1}--\ref{sec:exp3} do not
resolve cleanly into either of the two outcomes this audit set out to
distinguish. Experiment~\ref{sec:exp1} explains, rather than merely
documents, the original instability: it is substantially attributable
to the ratio-of-maximum construction of the original statistic, not to
an unexplained property of the estimator or the data, and the direct
mean-difference alternative is stable by one to two orders of magnitude
on the same cells. In that specific sense, the first of the two framed
outcomes partly holds: a change in \emph{statistic}, not estimator,
resolves most of the instability Experiment~\ref{sec:exp1} set out to
characterize. Experiment~\ref{sec:exp2} extends this partial positive
result to estimator choice at matched dimensionality: two of three
alternatives to DV/MINE are cross-seed, cross-run, and
cross-pairing stable on ESM-2 under the new statistic.

This gain does not survive the generalization test in
Experiment~\ref{sec:exp3}. Both estimators that stabilized at matched
dimensionality, run blind on synthetic channels at the dimensionalities of two additional models, fail
to recover a signal large enough to clear the strong-signal margin,
despite correct signal ordering and stable seed-level estimates --
stability under training variation is not the same property as
sufficient signal recovery under blind search. The oracle control is
what keeps this from collapsing into the second framed outcome
outright: because a critic given the true signal direction (given a one-dimensional input) recovers it at both dimensionalities, the Experiment~\ref{sec:exp3} failure is better
read as evidence against \emph{blind} estimation specifically at this
dimensionality, not as evidence about these models' representations, which the synthetic calibration does not use. The practical implication for a researcher asking an
order-resolved dependence question is accordingly narrower than either
framed outcome alone: estimator and statistic choice can resolve
training-related instability at a model and dimensionality already
matched to prior calibration, but neither currently licenses a blind, order-resolved claim at a higher representation dimensionality without the kind of oracle-style check -- or the full diagnostic protocol in
Section~\ref{sec:exp4} -- this audit required to tell a real signal
from a search failure.

\section{Limitations}

This audit is restricted to one biological landscape (four-site GB1),
one class of interaction-order question (peak-to-final retention across
model depth), and a bounded set of candidate estimators; a negative
result here does not establish that no estimator can ever succeed in
this regime, only that the ones tested here do not. Experiment~\ref{sec:exp3} uses synthetic channels matched to the models only in dimension and sample size; it
contains no ESM-1v or ProtT5 representations and so cannot speak to model-specific effects, and a dimension sweep that
would separate dimensionality from the oracle's knowledge advantage was not run. Real-data critic runs exist for ESM-2
only; the evidence from the other two models' own representations is the ridge probe, under the IID split. The
$\tau$ floor for DV/MINE reported here ($0.007634$, from the IID null) differs from the $0.006777$ recorded in some run
manifests, and we did not test whether this changes downstream statistics. The oracle control in Experiment 3
establishes that the signal is recoverable given its direction, but
does not identify what makes blind search specifically difficult at
this dimensionality; that mechanism is not characterized here. The
descriptive checks in Experiment 2 (Section~\ref{sec:exp2}) report
associations between instability and clip activation, interaction
order, and signal magnitude, but do not establish causal drivers of any
estimator's instability. Finally, the diagnostic protocol in
Section~\ref{sec:exp4} is derived entirely from what this single
landscape and estimator set revealed; its generality to other
biological systems or representation types has not been tested.

\section{Conclusion}

This audit set out to determine whether a specific instability in an
order-resolved dependence claim about protein language models was
explainable, fixable, and general, and arrives at a qualified answer to
each. The instability is largely explainable: it traces substantially
to a ratio-of-maximum statistic rather than to an unexplained property
of the estimator or the underlying data. It is fixable at the model and
dimensionality where it was first observed: two of three alternative
estimators, paired with a direct mean-difference statistic, are stable
across seeds, independent runs, and evaluation pairing on ESM-2. It is not,
however, general: that same fix does not transfer to the higher representation dimensionalities of two other models under blind search on a synthetic channel at this sample size, where even the stabilized estimators recover materially
less signal than an oracle shown the true direction can recover. The practical implication is that estimator and statistic
choice are necessary but not sufficient safeguards for this class of
claim -- a blind, order-resolved dependence claim on a new model or
representation should not be reported on the strength of a stable
estimate alone, without the kind of direct stability, pairing, and
oracle-style recoverability checks packaged in
Section~\ref{sec:exp4}, run and reported alongside the claim itself.

\clearpage
\bibliographystyle{apalike}
\bibliography{mi_reliability_audit_references}

\clearpage
\appendix

\section{Construction of the Direct, Pooled Mean-Difference Comparison}
\label{app:direct-comparison}

This appendix gives the full construction of the diagnostic referenced
in Section~\ref{sec:exp1}. It is included for completeness and
reproducibility, not as a proposed general-purpose estimator.

The original protocol's statistic is built by dividing each subset's
final-layer score by that subset's own peak score across all 31 layers,
then averaging these ratios within the higher- and lower-order groups.
This construction requires the estimator to correctly identify and
certify a \emph{maximum} over noisy per-layer estimates before any
comparison is made. Order statistics of this kind are known to inflate
apparent signal relative to the underlying per-observation noise, and
this inflation is more severe for a maximum taken over more values ---
here, the four interaction orders draw from 4, 6, 4, and 1 positional
subsets respectively, so this effect is not applied uniformly across
the quantity being compared.

The diagnostic reported in Section~\ref{sec:exp1} avoids this specific
construction. For each layer, it pools the raw per-subset scores
belonging to the higher- and lower-order groups directly (subject to
the same derangement-null floor $\tau$ used throughout this protocol,
applied per subset per layer rather than to a single peak value) and
computes the difference between the pooled group means directly:
$\mathrm{Gap} = \mathrm{mean}(\mathrm{LO\ scores}) -
\mathrm{mean}(\mathrm{HO\ scores})$, the quantity reported in
Table~\ref{tab:exp2}. This is a direct contrast of raw values, not a
rank-based or distribution-free test; it makes no claim about
stochastic dominance between the two groups' distributions, only
about the difference in their means.

A stable value under this statistic establishes that the pooled mean
contrast itself is reproducible across seeds, independent runs, and evaluation pairing. It does not, on its own, establish that
peak-to-final retention -- the original protocol's own quantity
(Section~\ref{sec:recap}) -- is similarly stable, since the two
statistics answer different questions rather than being
interchangeable, as Figure~\ref{fig:framework} already states.

\clearpage
\section{Full Strong-Signal-Margin Gate Data}
\label{sec:margin-gate-appendix}

Table~\ref{tab:margin-gate-full} reports every individual
replicate/gain comparison underlying Figure~\ref{fig:margin-gate} and
Experiment 3's calibration outcomes: three replicates times
two strong gains (1.0, 1.5) for each of DV/MINE, Bounded-DV, SMILE,
and Oracle, at both the $5{,}120$-D and $4{,}096$-D calibration channels -- 48 comparisons in total. Each
margin was independently recomputed from the real saved per-replicate
summaries rather than taken from the manifest's own stored value, and
each run's aggregate pass/fail outcome was then cross-checked against
its manifest; all 8 runs agree. A run passes the overall calibration
gate only if all six of its comparisons individually meet the $0.05$
margin requirement.

\begin{longtable}{llrrrrrrc}
\caption{Full strong-signal-margin gate data: every replicate/gain comparison underlying Figure~\ref{fig:margin-gate}, independently recomputed from the real per-replicate summaries (\texttt{replicate\_summary.csv}) on both models, for all four conditions. Margin $=$ estimate $-$ that replicate's own null estimate; shortfall $=$ 0.05 $-$ margin. Pass/fail uses the unrounded comparison estimate $\geq$ null estimate $+$ 0.05; displayed values are rounded to six decimal places. Oracle is a privileged diagnostic -- given the true signal direction rather than finding it blind -- not a fourth blind estimator. A run passes this margin gate only when all six of its comparisons pass; manifest agreement here concerns that gate specifically, not overall calibration.}
\label{tab:margin-gate-full}\\
\toprule
Model & Estimator & Rep. & Gain & Estimate & Null est. & Margin & Shortfall & \shortstack{Comparison\\passes} \\
\midrule
\endfirsthead
\multicolumn{9}{l}{\textit{(continued from previous page)}} \\
\toprule
Model & Estimator & Rep. & Gain & Estimate & Null est. & Margin & Shortfall & \shortstack{Comparison\\passes} \\
\midrule
\endhead
\midrule
\multicolumn{9}{r}{\textit{(continued on next page)}} \\
\endfoot
\bottomrule
\endlastfoot
ProtT5 & DV/MINE & 101 & 1.0 & 0.039377 & -0.000219 & 0.039596 & 0.010404 & No \\
ProtT5 & DV/MINE & 101 & 1.5 & 0.142699 & -0.000219 & 0.142918 & -0.092918 & Yes \\
ProtT5 & DV/MINE & 211 & 1.0 & 0.047836 & -0.000170 & 0.048006 & 0.001994 & No \\
ProtT5 & DV/MINE & 211 & 1.5 & 0.149701 & -0.000170 & 0.149871 & -0.099871 & Yes \\
ProtT5 & DV/MINE & 307 & 1.0 & 0.040352 & -0.000121 & 0.040474 & 0.009526 & No \\
ProtT5 & DV/MINE & 307 & 1.5 & 0.127060 & -0.000121 & 0.127181 & -0.077181 & Yes \\
\addlinespace
ProtT5 & Bounded-DV & 101 & 1.0 & 0.008879 & -0.000071 & 0.008950 & 0.041050 & No \\
ProtT5 & Bounded-DV & 101 & 1.5 & 0.156170 & -0.000071 & 0.156241 & -0.106241 & Yes \\
ProtT5 & Bounded-DV & 211 & 1.0 & 0.043047 & -0.000204 & 0.043251 & 0.006749 & No \\
ProtT5 & Bounded-DV & 211 & 1.5 & 0.166379 & -0.000204 & 0.166583 & -0.116583 & Yes \\
ProtT5 & Bounded-DV & 307 & 1.0 & 0.052522 & -0.000187 & 0.052709 & -0.002709 & Yes \\
ProtT5 & Bounded-DV & 307 & 1.5 & 0.187211 & -0.000187 & 0.187398 & -0.137398 & Yes \\
\addlinespace
ProtT5 & SMILE & 101 & 1.0 & 0.036006 & -0.000086 & 0.036092 & 0.013908 & No \\
ProtT5 & SMILE & 101 & 1.5 & 0.160795 & -0.000086 & 0.160881 & -0.110881 & Yes \\
ProtT5 & SMILE & 211 & 1.0 & 0.034154 & -0.000179 & 0.034332 & 0.015668 & No \\
ProtT5 & SMILE & 211 & 1.5 & 0.137534 & -0.000179 & 0.137713 & -0.087713 & Yes \\
ProtT5 & SMILE & 307 & 1.0 & 0.046996 & -0.001041 & 0.048038 & 0.001962 & No \\
ProtT5 & SMILE & 307 & 1.5 & 0.154145 & -0.001041 & 0.155186 & -0.105186 & Yes \\
\addlinespace
ProtT5 & Oracle & 101 & 1.0 & 0.294802 & -0.000567 & 0.295370 & -0.245370 & Yes \\
ProtT5 & Oracle & 101 & 1.5 & 0.566684 & -0.000567 & 0.567251 & -0.517251 & Yes \\
ProtT5 & Oracle & 211 & 1.0 & 0.353138 & 0.000030 & 0.353108 & -0.303108 & Yes \\
ProtT5 & Oracle & 211 & 1.5 & 0.622544 & 0.000030 & 0.622514 & -0.572514 & Yes \\
ProtT5 & Oracle & 307 & 1.0 & 0.333616 & 0.000632 & 0.332983 & -0.282983 & Yes \\
ProtT5 & Oracle & 307 & 1.5 & 0.556638 & 0.000632 & 0.556006 & -0.506006 & Yes \\
\midrule
ESM-1v & DV/MINE & 101 & 1.0 & 0.030679 & -0.000207 & 0.030885 & 0.019115 & No \\
ESM-1v & DV/MINE & 101 & 1.5 & 0.071104 & -0.000207 & 0.071311 & -0.021311 & Yes \\
ESM-1v & DV/MINE & 211 & 1.0 & 0.047299 & -0.000273 & 0.047572 & 0.002428 & No \\
ESM-1v & DV/MINE & 211 & 1.5 & 0.156491 & -0.000273 & 0.156764 & -0.106764 & Yes \\
ESM-1v & DV/MINE & 307 & 1.0 & 0.041112 & -0.000184 & 0.041296 & 0.008704 & No \\
ESM-1v & DV/MINE & 307 & 1.5 & 0.125818 & -0.000184 & 0.126002 & -0.076002 & Yes \\
\addlinespace
ESM-1v & Bounded-DV & 101 & 1.0 & 0.054314 & -0.000214 & 0.054528 & -0.004528 & Yes \\
ESM-1v & Bounded-DV & 101 & 1.5 & 0.084652 & -0.000214 & 0.084866 & -0.034866 & Yes \\
ESM-1v & Bounded-DV & 211 & 1.0 & 0.056969 & -0.000286 & 0.057255 & -0.007255 & Yes \\
ESM-1v & Bounded-DV & 211 & 1.5 & 0.181943 & -0.000286 & 0.182229 & -0.132229 & Yes \\
ESM-1v & Bounded-DV & 307 & 1.0 & 0.036093 & -0.000141 & 0.036235 & 0.013765 & No \\
ESM-1v & Bounded-DV & 307 & 1.5 & 0.124811 & -0.000141 & 0.124953 & -0.074953 & Yes \\
\addlinespace
ESM-1v & SMILE & 101 & 1.0 & 0.038104 & -0.000174 & 0.038279 & 0.011721 & No \\
ESM-1v & SMILE & 101 & 1.5 & 0.099624 & -0.000174 & 0.099799 & -0.049799 & Yes \\
ESM-1v & SMILE & 211 & 1.0 & 0.054204 & -0.000161 & 0.054365 & -0.004365 & Yes \\
ESM-1v & SMILE & 211 & 1.5 & 0.147765 & -0.000161 & 0.147926 & -0.097926 & Yes \\
ESM-1v & SMILE & 307 & 1.0 & 0.039050 & -0.000130 & 0.039180 & 0.010820 & No \\
ESM-1v & SMILE & 307 & 1.5 & 0.105714 & -0.000130 & 0.105844 & -0.055844 & Yes \\
\addlinespace
ESM-1v & Oracle & 101 & 1.0 & 0.385720 & 0.000993 & 0.384727 & -0.334727 & Yes \\
ESM-1v & Oracle & 101 & 1.5 & 0.529803 & 0.000993 & 0.528810 & -0.478810 & Yes \\
ESM-1v & Oracle & 211 & 1.0 & 0.420496 & -0.001739 & 0.422236 & -0.372236 & Yes \\
ESM-1v & Oracle & 211 & 1.5 & 0.579301 & -0.001739 & 0.581041 & -0.531041 & Yes \\
ESM-1v & Oracle & 307 & 1.0 & 0.317940 & -0.001871 & 0.319811 & -0.269811 & Yes \\
ESM-1v & Oracle & 307 & 1.5 & 0.540607 & -0.001871 & 0.542478 & -0.492478 & Yes \\
\end{longtable}
\section{Evaluation-Pairing Construction}
\label{app:pairing}

\begin{figure}[htbp]
    \centering
    \includegraphics[width=\linewidth]{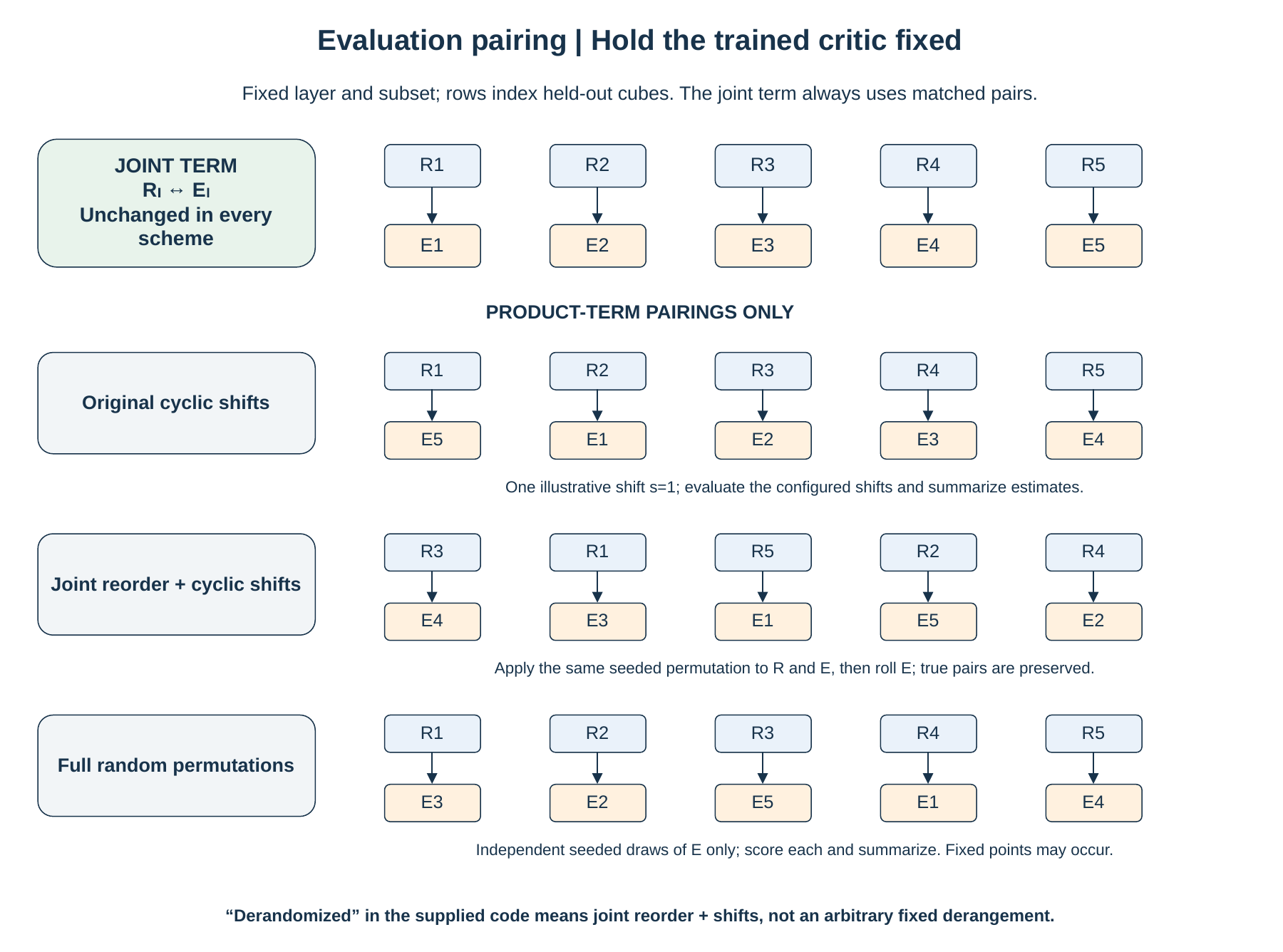}
    \caption{Evaluation-pairing schemes, holding the trained critic
    fixed. Rows index held-out cubes under a fixed layer and subset.
    The joint term $R_i \leftrightarrow E_i$ is unchanged across every
    scheme. Product-term pairings only: original cyclic shifts (one
    illustrative shift $s{=}1$ shown; the configured shifts are
    evaluated and summarized); joint reorder + cyclic shifts (the same
    seeded permutation applied to $R$ and $E$, then $E$ is rolled --
    true pairs are preserved under the permutation, not the roll);
    full random permutations (independent seeded draws of $E$ only,
    scored and summarized -- fixed points may occur, as shown for
    R2$\to$E2).}
    \label{fig:pairing}
\end{figure}

Figure~\ref{fig:pairing} illustrates the four evaluation-pairing
schemes referenced in Section~\ref{sec:exp4-pairing}: the joint term
always uses matched pairs regardless of scheme; the three product-term
pairing schemes differ in how the negative (mismatched) pairs are
constructed. ``Derandomized'' specifically means a joint reorder of
both $R$ and $E$ followed by a cyclic shift of $E$, which preserves no
true pairs by construction; this is distinct from a full random
permutation, which can retain fixed points (a drawn element mapped to
itself) with nonzero probability, as illustrated in the figure's final
row.

\FloatBarrier

\section{Order-Resolved Depth Curves}
\label{app:depth-curves}

Figure~\ref{fig:depth-curves} and Figure~\ref{fig:depth-curves-by-order}
report the full per-order, per-layer grid referenced in
Section~\ref{sec:exp2}, extending the pooled, single-layer contrast
already reported there (Table~\ref{tab:exp2}) across the full depth.

\begin{figure}[htbp]
\centering
\includegraphics[width=\linewidth]{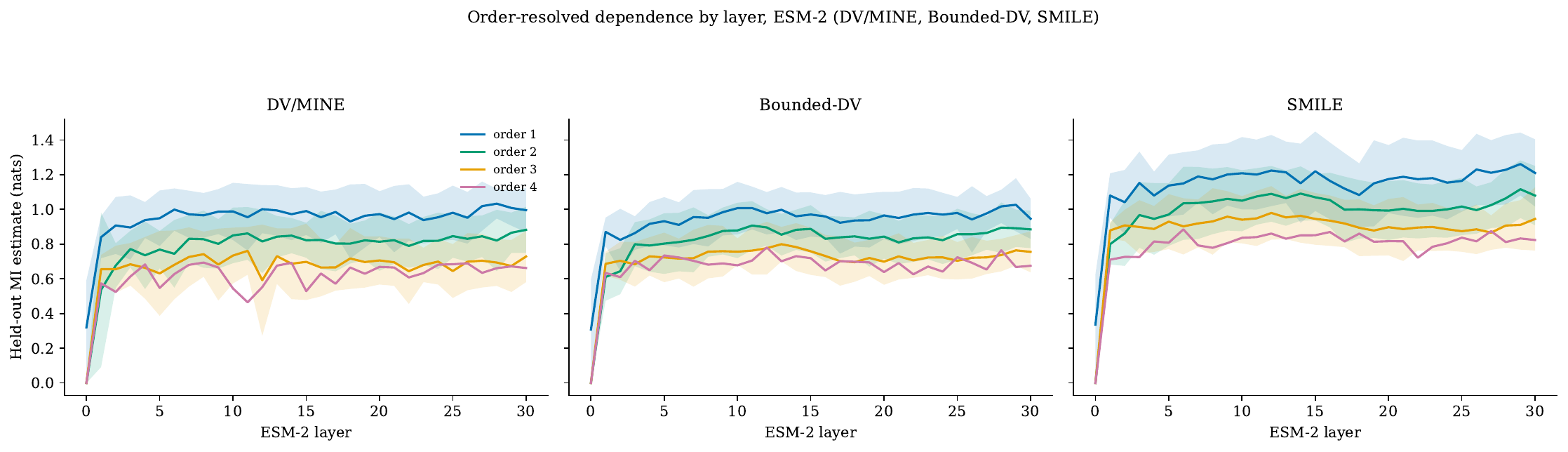}
\input{figures/depth_curves_caption.tex}
\end{figure}

\begin{figure}[htbp]
\centering
\includegraphics[width=\linewidth]{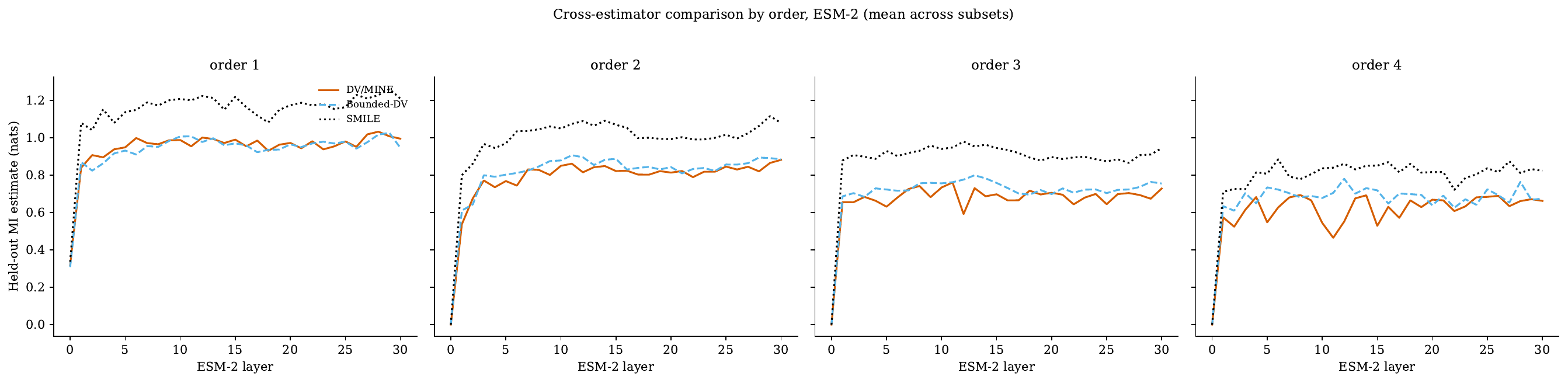}
\input{figures/depth_curves_by_order_caption.tex}
\end{figure}
\FloatBarrier
\section{ORBIT-Synth Validation Suite}
\label{app:orbit-synth}

The source protocol's neural dependence estimator is treated as a
measurement instrument requiring validation before any biological
interpretation. This appendix summarizes the five checks referenced
in Section~\ref{sec:exp2}'s ORBIT-Synth validation; full detail is in
\citet{rahimimovassagh2026orbitfmib}, Appendices H.1--H.5.

\paragraph{Known-ground-truth recovery.} The complete
representation--Walsh--dependence pipeline was first validated on
ORBIT-Synth, a synthetic benchmark where the true interaction order
and the location of imposed information-loss events are known in
advance. The benchmark tests whether the pipeline can recover known
first-, second-, third-, and fourth-order interaction signals; produce
low dependence scores for absent or off-target interaction orders;
localize deliberately imposed order-specific information-loss events;
preserve this behavior across the pre-specified critic seeds; and
reproduce it on held-out samples rather than only on critic-training
data. All pre-specified acceptance criteria were met before any
biological interpretation was attempted.

\paragraph{Dimension-specific absolute-MI calibration.} Passing a
low-dimensional synthetic test does not guarantee accurate estimation
at the much higher 2{,}560-dimensional representation size used in the
real biological analysis ($4 \times 640$, the ESM-2 mutation-site
representation). A separate synthetic calibration was therefore run at
this exact dimensionality. The estimator correctly represented the
null and preserved the expected ordering of synthetic dependence
strength, but substantially underestimated moderate and strong
analytic mutual information at this scale. This failed
absolute-magnitude criterion was not bypassed: scores are accordingly
never interpreted as calibrated absolute mutual information in nats,
only as comparative signal, consistent with how $D_{\ell,S}$ is used
throughout this paper.

\paragraph{Relative-retention calibration.} A separate synthetic
trajectory, constructed at the same 2{,}560-dimensional scale with a
common rising information profile and four order-analogue final
states of known peak-to-final retention, tested whether the same fixed
estimator and pre-specified critic seeds could recover the correct
relative ordering, rather than absolute magnitude. The estimator
recovered the expected retention ordering and the expected negative
higher-order-minus-lower-order contrast; every pre-specified hard
gate passed, including null behavior, seed stability, peak
localization, signal separation, and contrast direction. This
establishes suitability for comparative, order-resolved retention
analysis specifically, not for absolute-MI recovery.

\paragraph{Hypothesis-adversarial order-pattern validation.} To confirm
the procedure was not simply calibrated to reproduce the expected
higher-order-loss pattern, four adversarial scenarios were tested:
higher-order loss, higher-order advantage, flat strong retention, and
flat moderate retention. The two directional scenarios recovered the
expected contrast direction. The two flat scenarios are worth a
specific methodological note: the original flat-scenario test used an
exact four-label permutation test whose minimum two-sided $p$-value is
structurally $1/3$, meaning it could never reject at the $0.05$
threshold regardless of the underlying data -- an uninformative test
by construction, not a genuine pass. This was corrected with a
multi-replicate Monte Carlo permutation test (20 independent replicate
instances per scenario, resolution $\approx 5 \times 10^{-6}$), under
which both flat scenarios showed no detectable deviation from flatness
($p = 0.949$ and $p = 0.929$ respectively) -- genuine evidence against
spurious order selectivity, not an artifact of an underpowered test.

\paragraph{Fixed-subset shortcut removal.} This check is the direct
motivation for the fixed-subset protocol this entire paper audits. An
earlier estimator that pooled different positional subsets of the same
interaction order failed an adversarial cube-derangement control:
after derangement, apparent dependence remained, because positional
subset identity itself continued to be matched between the
representation and target distributions. This revealed a shortcut
pathway -- different positional subsets can have systematically
different empirical distributions of epistatic coefficients, letting
a pooled critic exploit subset identity rather than the genuine,
matched cube-specific representation--target relationship. The
protocol was reformulated so that each critic is conditioned on one
fixed positional subset $S$; under this correction, cube-deranged null
scores collapse near the neural-estimator floor. This is why $S$ is
held fixed within every critic throughout this paper
(Section~\ref{sec:recap}).

\end{document}